\documentclass[conference]{IEEEtran}

\usepackage{soul}
\usepackage{graphicx}
\usepackage{caption}
\usepackage{multirow,multicol}
\usepackage{xcolor}
\usepackage{cite}   
\usepackage[hidelinks]{hyperref}
\def\BibTeX{{\rm B\kern-.05em{\sc i\kern-.025em b}\kern-.08em
    T\kern-.1667em\lower.7ex\hbox{E}\kern-.125emX}}

\title{Distilled Large Language Model in Confidential Computing Environment for System-on-Chip Design}

\author{
    \IEEEauthorblockN{Ben Dong, Hui Feng, Qian Wang} 
    \IEEEauthorblockA{
        University of California, Merced\\
        \{cdong12, hfeng9, qianwang\}@ucmerced.edu}
}
\begin{document}

\maketitle
\begin{abstract}
Large Language Models (LLMs) are increasingly used in circuit design tasks and have typically undergone multiple rounds of training. Both the trained models and their associated training data are considered confidential intellectual property (IP) and must be protected from exposure. Confidential Computing offers a promising solution to protect data and models through Trusted Execution Environments (TEEs). However, existing TEE implementations are not designed to support the resource-intensive nature of LLMs efficiently. In this work, we first present a comprehensive evaluation of the LLMs within a TEE-enabled confidential computing environment, specifically utilizing Intel Trust Domain Extensions (TDX). 
We constructed experiments on three environments: TEE-based, CPU-only, and CPU-GPU hybrid implementations, and evaluated their performance in terms of tokens per second.

Our first observation is that distilled models, i.e., DeepSeek, surpass other models in performance due to their smaller parameters, making them suitable for resource-constrained devices. Also, in the quantized models such as 4-bit quantization (Q4) and 8-bit quantization (Q8), we observed a performance gain of up to 3× compared to FP16 models. Our findings indicate that for fewer parameter sets, such as DeepSeek-r1-1.5B, the TDX implementation outperforms the CPU version in executing computations within a secure environment. We further validate the results using a testbench designed for SoC design tasks.
These validations demonstrate the potential of efficiently deploying lightweight LLMs on resource-constrained systems for semiconductor CAD applications.

\end{abstract}





\maketitle

\section{Introduction}
Large language models (LLMs) show significant capabilities in processing multimodal information, performing complex reasoning, and generating code, making them valuable tools for Computer-aided Design (CAD) tasks in System-on-Chip (SoC) designs. However, the use of LLMs in circuit design inevitably involves providing sensitive information, such as RTL designs or circuit specifications. This raises concerns about potential data breaches and the exposure of proprietary specifications through reverse data inference attacks. Even when smaller LLMs such as DeepSeek-R1 (1.5 billion parameters) or Llama 3.2 (1 billion parameters) are deployed locally to preserve data privacy, sensitive information may still be vulnerable to memory-snooping or side-channel attacks. Confidential computing through Trusted Execution Environments (TEEs) provides a promising solution to secure both data and model parameters.

TEE provides a secure enclave for computation and safeguards both data and models in cloud computing environments. Prior research has leveraged TEEs, such as Intel’s Software Guard Extensions (SGX) \cite{intel_sgx_whitepaper}, to secure ML workloads by isolating sensitive computations within secure enclaves \cite{narra2019privacy}. While SGX offers strong protection, its limited memory capacity (approximately 1 GB) and complex interface pose challenges for large-scale ML applications \cite{shen2022soter,sun2023shadownet}. To overcome these limitations, we use Intel’s Trust Domain Extensions (TDX) \cite{intel_tdx_whitepaper}, which introduces secure Virtual Machines (VMs) that can process larger ML models within the secure enclaves. 

Even though TDX significantly expands memory capacity compared to its predecessors, deploying advanced LLMs with large parameter counts such as GPT \cite{brown2020language} and Gemini \cite{deepmind2023gemini} in a confidential computing environment presents additional challenges.
To begin with, these models are relatively large, typically starting from at least 1 billion parameters and reaching over 100 billion or more. Even hosting LLMs locally faces significant difficulties due to memory constraints \cite{lin2025understanding}. In the SoC design domain, LLMs are increasingly used to process sensitive data such as netlists, design constraints, and proprietary specifications, further elevating the need for secure execution. In addition, LLMs used in hardware design are often fine-tuned with proprietary datasets and specifications \cite{fu2023llm4sechw,tarek2024socurellm}, which must be evaluated within confidential environments to ensure data security. Furthermore, both training and inference require substantial computational resources, and ensuring secure, optimized deployment in TEE-based settings like TDX involves resolving key constraints. 

Among these LLMs, DeepSeek stands out as an advanced model optimized for efficient resource utilization \cite{bi2024deepseek}. Its key distinction lies in its ability to conserve computational resources while maintaining strong reasoning capabilities. For example, the distilled version of DeepSeek significantly reduces model size and memory footprint, enabling effective performance. This efficiency is particularly valuable in secure settings such as TDX, where both performance and data security are critical \cite{dong2025evaluating}. 

This study explores the adaptability of LLMs in secure computing environments, with a particular emphasis on their use in SoC designs. We compare performance across TEE (confidential computing), CPU-only, and hybrid CPU-GPU implementations to assess their computational efficiency and suitability for confidential computing. This work aims to identify optimal models and deployment strategies for LLMs on resource-constrained devices, with a focus on maintaining robust data security.

The main contributions of this paper are summarized as follows:
\begin{itemize}
    \item We compare the performance across TEE-based, CPU-only, and CPU-GPU implementations, identifying key performance bottlenecks and trade-offs between security and computational efficiency. These findings offer practical insights for confidential computing vendors aiming to address scalability challenges in secure AI workloads.
    \item To the best of our knowledge, this is the first evaluation of a distilled LLM within a TEE. We provide a comprehensive analysis of its performance and behavior in a confidential computing environment.
    \item We evaluate lightweight LLMs with fewer than 8 billion parameters running in TEEs for use in SoC design. The results show that their performance in TEEs exceeds that of traditional CPU-only execution.
    \item We conduct a detailed benchmarking analysis of different quantization levels of LLMs and assess their efficiency for deployment in confidential computing settings.

\end{itemize}

\section{Backgrounds and State-of-the-art}

\subsection{Trust Execution Environment}
Trusted Execution Environment is a hardware-based security technology designed to protect data and programs from malicious outsiders by introducing an isolation mechanism known as the Trust Domain (TD).
This ensures that sensitive computations performed in the secure enclave are shielded from unauthorized access or tampering. Modern commercial CPUs implement different forms of isolation. For instance, Intel SGX \cite{intel_sgx_whitepaper} creates process-level isolation, which enables the secure execution of sensitive workloads within protected enclaves. Intel TDX \cite{intel_tdx_whitepaper} offers a more advanced form of memory isolation at the virtual machine (VM) level, allowing an entire VM to operate securely. Similarly, AMD Secure Encrypted Virtualization (SEV) \cite{amd_sev_whitepaper} provides hardware-based memory encryption for virtual machines. ARM TrustZone introduces a dual-world execution model, where the secure world runs trusted applications separately from the normal world, ensuring secure execution for tasks such as cryptographic key management, secure boot, and mobile security \cite{arm_trustzone}.

One of the key features of TEE is to protect data in use by encrypting all communications between the Trust Domain and external environments. Data entering or leaving the TD remains encrypted, thereby mitigating risks such as memory-snooping attacks and side-channel exploits. Additionally,  the TD allocates a private memory space where sensitive data can be processed securely, without exposure to the untrusted host system. This capability is particularly crucial for confidential computing applications, including secure AI inference, financial transactions, and healthcare data processing, where both data privacy and computational performance are essential.

While TEEs offer strong security guarantees, they also introduce notable performance overhead due to the mechanisms required to maintain data isolation. A primary limitation is the increased computational latency resulting from memory encryption and secure data handling within the TD. Since all data exchanges between the secure enclave and external components must undergo encryption and decryption, this process incurs significant processing overhead. Moreover, TEEs often restrict direct access to hardware accelerators such as GPUs \cite{lin2023drgpum}. For example, Intel TDX does not yet fully support GPU acceleration within the secure enclave, forcing AI models to rely on CPU-based execution, which is significantly slower than GPU-accelerated implementations \cite{11014474}. Furthermore, memory limitations in TEEs, particularly in Intel SGX, can further degrade performance due to frequent memory swapping and paging \cite{lin2025forest}. These constraints create a trade-off between security and computational efficiency. To address this, optimization strategies such as model quantization, workload partitioning, and hybrid CPU-GPU confidential computing should be explored to mitigate performance bottlenecks while maintaining robust security guarantees.
\begin{figure}[htb]
    \centering
    \captionsetup{font=small} 
    \includegraphics[width=1.0\linewidth]{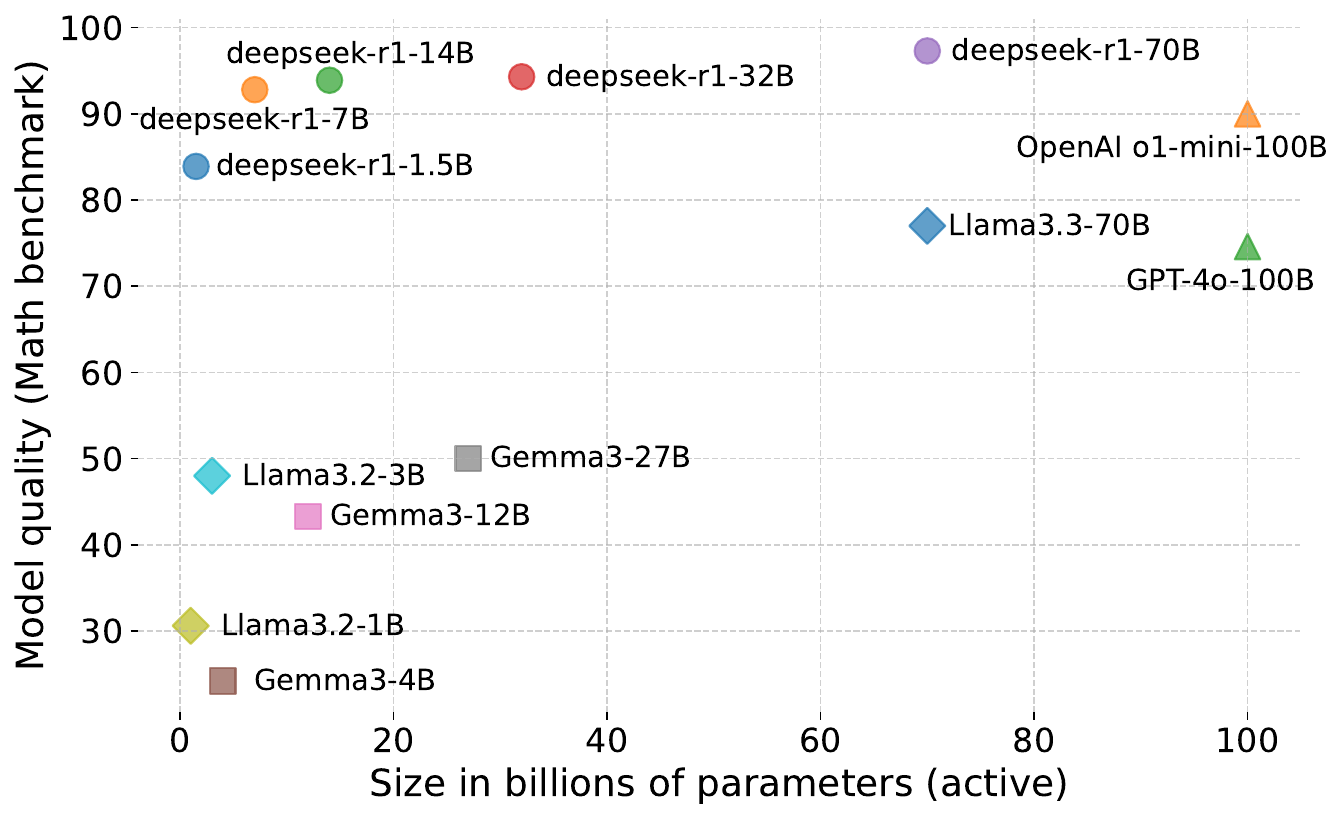}
    \caption{Comparison of math performance vs. model size, Larger models generally show improved performance, but variations exist across architectures.}
    \label{fig:perf_model}
\end{figure}
\subsection{Distilled Large Language Models}
A distilled LLM is a scaled-down version that retains much of the original capability of a larger and more sophisticated model while significantly reducing its size, latency, and computational demands. The distillation process involves training a smaller model, commonly referred to as the student model, to mimic the behavior of a larger teacher model by transferring knowledge through techniques such as intermediate layer distillation, data augmentation, and label enrichment. Distilled models provide faster inference speeds and require less memory, making them well-suited for resource-constrained confidential computing environments where computational efficiency is crucial.

A prime example of distilled LLM is the DeepSeek-r1. Its distilled variants, such as DeepSeek-r1-qwen 1.5B, demonstrate substantial efficiency improvements by reducing model size and memory footprint without compromising accuracy. Figure \ref{fig:perf_model} illustrates the performance of DeepSeek across various model sizes on the Math500 dataset. The results indicate that DeepSeek maintains robust performance even when distilled to smaller parameter sizes, such as 1.5B and 7B. These findings suggest that model distillation effectively preserves DeepSeek’s core capabilities while enabling faster inference and lower resource consumption without significant degradation of accuracy. Such characteristics make distilled models particularly well-suited for confidential computing environments, where deployability, efficiency, and security are prioritized.

\subsection{LLM usage in System-on-Chip Design}
The increasing complexity of semiconductor designs, particularly system-on-chip (SoC) architectures, has introduced significant challenges in verification, debugging, and security assurance. Recent advancements in domain-specific LLMs offer promising capabilities to address these issues. By leveraging their ability to process and understand hardware description languages, LLMs can analyze design files, detect bugs, and even generate security policies. SoCureLLM introduces a comprehensive framework that partitions large hardware designs into manageable segments and uses a structured prompting approach to identify vulnerabilities and generate security policies \cite{tarek2024socurellm}. LLM4SecHW employs fine-tuning of medium-sized LLMs on curated datasets from version control histories of open-source hardware projects, enabling accurate bug localization and repair \cite{fu2023llm4sechw}. These approaches highlight a shift toward LLM-driven design automation in semiconductor engineering.

Building upon these efforts, other frameworks have emerged to address specific challenges in hardware security. For instance, MARVEL introduces a multi-agent LLM framework that mimics the cognitive process of a designer to identify security vulnerabilities in RTL code, integrating formal tools, linters, and simulation tests to enhance detection accuracy \cite{collini2025marvel}. Similarly, SPICED leverages LLMs to detect syntactical bugs and analog Trojans in circuit netlists without requiring hardware modifications, achieving high true positive rates in identifying Trojan-impacted nodes \cite{chaudhuri2024spiced}. Additionally, ThreatLens employs LLM-driven multi-agent systems to automate security threat modeling and test plan generation, reducing manual verification efforts and enhancing coverage \cite{saha2025threatlens}. All of these proposed models require fine-tuning LLMs using domain-specific requirements or benchmarks, most of which involve confidential models and datasets that must be protected from external exposure.

\section{Deploying LLMs on TDX (TEE)}

\subsection{TEE Configuration} 
\begin{figure}[htb]
    \centering
    \captionsetup{font=small} 
    \includegraphics[width=1.0\linewidth]{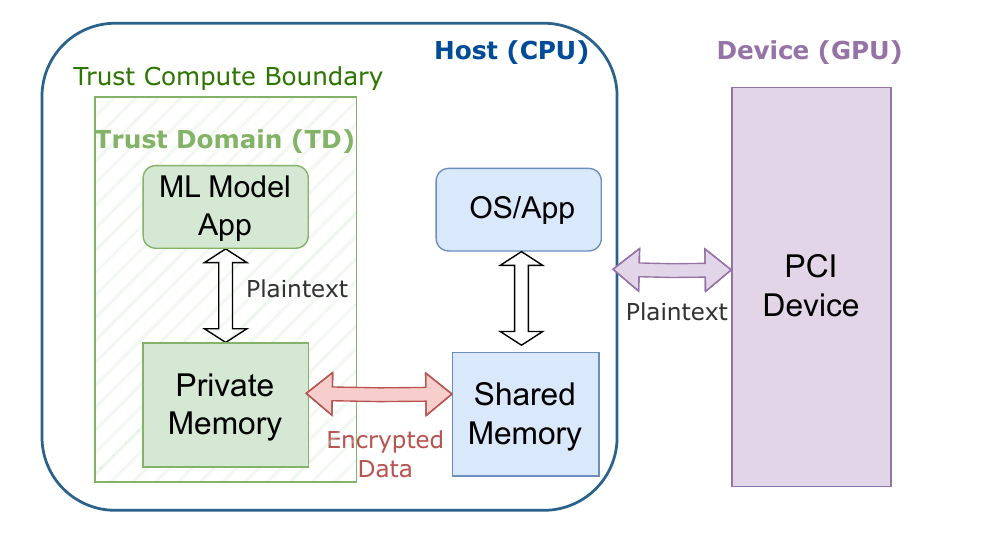}
    \vspace{-5mm}
    \caption{Trust Environment Settings with Data I/O to Shared memory and GPU.}
    \label{fig:tdx} 
\end{figure}

TEE-based confidential computing offers a robust solution by ensuring that all data and workloads are executed within a secure enclave, where input and output communications are encrypted, as shown in Figure \ref{fig:tdx}. This figure illustrates the settings on Intel TDX, where VM containers establish a trust compute domain to process applications in the secured trusted enclave.  A dedicated private memory region is allocated exclusively to the confidential execution environment, making memory capacity a critical constraint when deploying LLMs.  All data exchanges between the enclave and shared memory are encrypted and decrypted to preserve confidentiality.  Furthermore, the system interfaces with a PCIe-connected device to facilitate collaborative processing between the CPU and GPU, thus improving workload distribution efficiency. However, the communication between the CPU host and the GPU device currently occurs in plaintext, lacking adequate safeguards for protecting confidential data associated with SoC design processes. While encryption could be implemented at the PCIe protocol level, this paper assumes that lightweight distilled models are kept within the secure enclave, eliminating the immediate need for encryption. Nevertheless, future solutions could extend the trust domain to include PCIe-connected devices for larger models to enhance security while enabling GPU acceleration.


To evaluate the effectiveness of TEE settings, we implemented and tested three operational modes. The first mode, referred to as the TDX mode, executes all LLM computations entirely within the trust domain, using private memory to store both the model and the data. When sufficient memory is available and a smaller LLM model is used, performance in this environment can match or even exceed that of traditional setups. The second mode (CPU-only) stores the model and data in the system’s main memory and performs all computations directly on the host operating system, without the security protections offered by a TEE. The third mode follows a hybrid CPU-GPU configuration, in which the GPU handles computation while the CPU manages scheduling. Although this configuration achieves the highest performance among the three, it lacks confidentiality guarantees, as all data is processed in plaintext outside the trust domain.


\subsection{Model Selection} 
We evaluated the DeepSeek R1 model in three different model sizes: 1.5 billion, 7 billion, and 14 billion parameters. DeepSeek R1 is a distilled language model that demonstrates strong capabilities in reasoning, logic, and structured problem solving, making it suitable for technical domains such as programming, mathematics, and hardware design (e.g., RTL design). We also evaluated other lightweight models, such as Llama and Gemma, with fewer than 7B parameters in the TDX environment. The LLaMA series consists of open-weight models designed for efficiency and fine-tuning, offering strong performance per parameter along with broad support for customization, privacy, and deployment. Gemma models are built for responsible AI development, optimized for both on-device and cloud use, featuring robust safety measures and seamless integration into Google's ML ecosystem. 

DeepSeek-R1, LLaMA, and Gemma offer lightweight model variants, each with no more than 14 billion parameters. Their relatively small size enables efficient deployment on local machines with limited computational and memory resources, thereby providing a practical foundation for evaluating model performance in TDX, CPU-only, and CPU-GPU configurations. While LLMs show strong capabilities across various complex tasks, smaller models are more economical to develop and can be effectively fine-tuned for domain-specific applications such as SoC design. When provided with sufficient high-quality labeled data, these small models can achieve performance comparable to that of larger models in specific tasks \cite{qin2023chatgpt, zhong2023can}. To further support our evaluation of LLMs in CAD design tasks, we analyzed the KSU HWSEC dataset, which includes the Qwen model (the original source model before distillation in DeepSeek R1) and a fine-tuned Llama model used for RTL generation and bug detection. 

\subsection{Quantization for Performance}
Post-training quantization is a neural network compression technique that reduces numerical precision by converting floating-point values to integers for model weights and activations. Model quantization can reduce model size, accelerate inference, and lower memory and computational costs without significantly sacrificing accuracy. \cite{zhao2023survey} evaluates the performance of quantized Llama-7B and Llama-13B models at three precision levels: 4-bit, 8-bit, and 16-bit. The results indicate that 4-bit and 8-bit weight quantization is close to the performance of 16-bit models while significantly reducing memory usage. It is beneficial for deploying neural networks on resource-constrained devices such as personal computers and embedded systems.   In our experiments, we benchmarked and evaluated the Q4 and Q8 versions of DeepSeek within the TDX environment, analyzing their impact on overall inference performance.

\section{Experimental Setting}
\subsection{Evaluation Platform}
The evaluation was conducted using a high-performance host machine equipped with two Intel Xeon Gold 6530 CPUs, each featuring 32 cores and 512 GB of DDR5 memory operating at 4800 MHz. The testing environment was deployed within an Intel TDX virtual machine, configured according to Canonical Ubuntu’s official installation guide. To benchmark the performance of LLM inference, a container was launched from Ollama’s official image with CPU and memory constraints simulating the TDX VM configuration. A comparative evaluation was performed using Docker containers on the host CPU. Additionally, a separate container is created with GPU access using NVIDIA’s container toolkit to assess the performance differential between CPU-based execution and GPU acceleration. All performance data was gathered using Ollama’s built-in logging mechanisms, ensuring consistency across test conditions. 


\begin{table}[h]
    \centering
    \captionsetup{font=small} 
    \caption{Experiment Evaluation Configuration}
    \begin{tabular}{l|p{5cm}}
        \hline
        \textbf{Configuration} & \textbf{Description} \\
        \hline
        CPU Only &Runs on the host machine without GPU acceleration enabled. \\
        \hline
        TDX & Runs in a TD container with 62 CPU cores and 510GB DRAM. \\
        \hline
        GPU-CPU & Runs with GPU acceleration enabled. \\
        \hline
    \end{tabular}
    \label{tab:exp-set}\vspace{-8pt}
\end{table}

\subsection{Inference Workload} 
LLM inference within a TEE involves executing a trained machine learning model to generate outputs based on user-provided prompts.  In this benchmark, we evaluate different configurations where either a CPU or a CPU-GPU combination handles the workload, as shown in Table \ref{tab:exp-set}. Under typical LLM inference settings, the CPU manages data loading, pre-processing, and post-processing, while the GPU is responsible for accelerating core computations.  However, in confidential computing mode, the enclave CPU takes on additional responsibilities for secure data processing to maintain model integrity and data privacy. The inference workflow begins with loading the trained machine-learning model into the memory allocated to the Trusted Domain (TD). Once inside the TD, computations occur in an isolated environment without direct communication with the untrusted external environment. This isolation guarantees that both sensitive data and model parameters remain protected from potential security threats.

We use the K-state HWSec dataset \cite{fu2024generalize} to benchmark the performance of LLM models in different settings. It is composed of fine-tuned LLMs based on open-source models such as LLaMA, DeepSeek, and Qwen, tailored for specific tasks including Verilog bug detection, RTL code generation, and related applications.
\section{Result Analysis} 

\begin{figure*}[htb]
    \centering
    \captionsetup{font=small} 
    \includegraphics[width=1.0\textwidth]{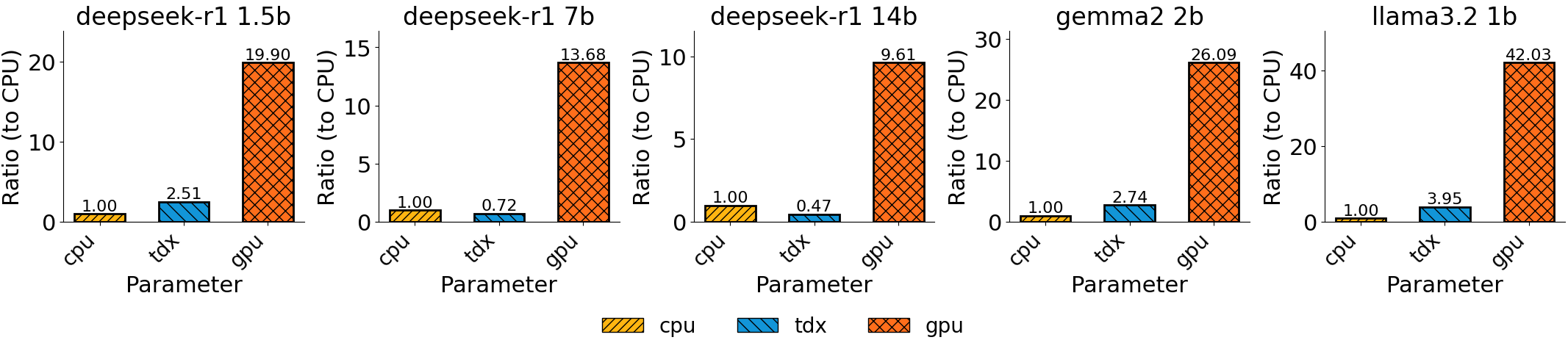}
    \caption{Performance of different models in different testing environments, ratioed to the CPU performance.}
    \vspace{-3mm}
    \label{fig:perf_tdx}
\end{figure*}

\subsection{General Performance of LLM on TDX}
We first employed DeepSeek R1 distilled models across three different parameter configurations: 1.5B, 7B, and 14B. Additionally, we tested Llama3.2 (1B and 3B) and Gemma 2 (2B) to provide a broader perspective of performance across different models. 
The results from Figure \ref{fig:perf_tdx} highlight the performance differences between Docker-based CPU execution and TDX settings, suggesting that TDX's optimizations for secure computing may also enhance CPU performance in certain scenarios. For example, in CPU-only configurations (as shown in Table \ref{tab:model_performance_gpu}), the TDX environment achieves the highest evaluation rate for the smallest model (1.5B), reaching approximately 25.67 tokens/s—more than twice the performance observed in CPU-only tests. This significant improvement could be attributed to TDX's optimized CPU ISA configuration, which minimizes inefficiencies. This advantage also applies to other smaller models, such as Gemma2-2B and Llama3.2 1B-3B. However, as the size of the model increases, the performance gap narrows, with both TDX and Docker CPU-based environments showing significant slowdowns for the 14B model, where TDX became slower than the CPU-only setting. This suggests that while TDX can leverage high-core configurations effectively for smaller models, larger models demand higher memory bandwidth and computational resources, limiting the advantage of TDX's CPU optimizations. Among the models compared, DeepSeek R1 demonstrates the best performance on CPU platforms.

Figure \ref{fig:perf_tdx} also shows that GPU acceleration drastically improves inference speed across all models. On average, GPU-based inference achieves approximately more than 20x the speed of CPU execution and 9× the speed of TDX execution for the smaller models, such as deepseek-1.5B, gemma2-2B, and llama3.2-3B. The 7B and 14B models also maintain high performance, achieving approximately 10-13 times performance gain, respectively. These results highlight the crucial role of GPU acceleration in the efficient deployment of LLM. However, current TDX implementations do not fully support GPU utilization within the secure enclave. Future research should focus on integrating GPU acceleration within TDX while preserving its security guarantees, enabling confidential AI inference to achieve both high security and computational efficiency.


\subsection{Evaluation Lightweight LLM on TDX}

\begin{table}[h]
\captionsetup{font=small} 
\centering
\caption{Comparison of Performance by Lightweight Model (Tokens/s)}
\begin{tabular}{l|c|c|c|c|c}
\hline
\multirow{2}{*}{\textbf{Model-Size}} & \multicolumn{3}{c|}{\textbf{Performance (tokens/s)}} & \multicolumn{2}{c}{\textbf{Comp. Ratio}} \\ 
 & \textbf{TDX} & \textbf{CPU only} & \textbf{CPU-GPU} & \textbf{$\frac{GPU}{TDX}$} & \textbf{$\frac{CPU}{TDX}$} \\ \hline
deepseek-1.5B & 25.67  & 10.25  & 202.88 & 7.9 & 0.4 \\ 
Llama3.2-1B & 22.9  & 7.11  & 243.24 & 10.6 & 0.31 \\ 
Llama3.2-3B & 12.83  & 2.71  & 176.09 & 13.72 & 0.21 \\ 
Gemma 2-2B & 14.81  & 4.29  & 140.81 & 9.51 & 0.29 \\ 
 deepseek-7B  & 6.42 & 8.53  & 117.02  & 18.2 & 1.3 \\ 
\hline
\end{tabular}
\label{tab:model_performance_gpu}
\end{table}

We further evaluate the advantages of leveraging Intel TDX by analyzing the trade-offs between security and computational performance when deploying DeepSeek R1. Our findings indicate that for lightweight LLMs (e.g., DeepSeek 1.5B), TDX outperforms CPU-only configurations and exhibits minimal performance degradation compared to CPU-GPU setups. Based on these observations, we extend our evaluation to additional lightweight models to further validate the effectiveness of TDX.

The results from Table \ref{tab:model_performance_gpu} highlight the significant performance differences between TDX, CPU only, and CPU-GPU settings. Comparing the CPU-only configurations, the TDX environment achieves the highest evaluation rate for the lightweight models ($<$3B), delivering more than twice the tokens per second. This performance advantage is consistent across most of the SOTA lightweight LLM models. The improvement can likely be attributed to TDX's optimized CPU execution, which reduces software-induced inefficiencies. However, as model size increases, the performance gap narrows. For the 7B model, both TDX and Docker CPU-based environments experience notable slowdowns, with the evaluation rate of the CPU-only configuration surpassing that of the TDX version. This suggests that while TDX effectively leverages a higher number of cores for smaller models, larger models usually require more memory bandwidth and computational resources, limiting the benefits of TDX's CPU optimizations. 

Additionally, Table \ref{tab:model_performance_gpu} shows that GPU acceleration improves inference speed by approximately 8 to 14 times, which is notably lower than the performance gains observed for larger models. This indicates that for lightweight inference models, if speed reduction is acceptable, a TEE-only approach can serve as a viable trade-off between performance and security.

\subsection{Evaluation on HWSeC LLM Models}
We further evaluate the performance of three HWSeC large language models—\texttt{Deepseek\_SFT\_RL\_Text\_AST}, \texttt{poison\_66-3B}, and \texttt{poison\_66-1B}—which are domain-adapted to hardware and security tasks. These models are fine-tuned from general-purpose bases such as DeepSeek and LLaMA, with extended context capabilities and curated datasets that emphasize chip design, hardware code generation, graph partitioning, security constraint creation, and RTL trojan insertion. The \texttt{Deepseek\_SFT\_RL\_Text\_AST} model specializes in code understanding and generation with an emphasis on abstract syntax tree (AST) context, while the \texttt{poison\_66} series targets hardware security analysis, long-context reasoning, and security-critical design modifications.

The results shown in Table~\ref{tab:model_hwsec} confirm that these domain-specific HWSeC models achieve improved accuracy and usability in chip design and verification benchmarks, outperforming general LLMs on hardware-centric tasks. While CPU-GPU deployments offer strong inference performance, they do not inherently provide model or data confidentiality. Notably, TDX-based secure enclaves deliver enhanced security protections and, for all HWSeC models,  exceeded raw CPU-only performance. This demonstrates that specialized, security-focused language models can be efficiently and securely deployed in practical hardware design and verification workflows.

\begin{table}[h]
\captionsetup{font=small} 
\centering
\caption{Comparison of Performance by HWSeC LLM Models.}
\resizebox{\columnwidth}{!}{ 
\begin{tabular}{c|c|c|c|c|c|c}
\hline
\multirow{2}{*}{\textbf{Model}} & \multirow{2}{*}{\textbf{Size}}  & \multicolumn{3}{c|}{\textbf{Performance (tokens/s)}} & \multicolumn{2}{c}{\textbf{Comp. Ratio}} \\ \cline{3-7}
 & & \textbf{GPU-CPU} & \textbf{CPU only} & \textbf{TDX} & \textbf{$\frac{GPU}{TDX}$} & \textbf{$\frac{CPU}{TDX}$} \\ \hline
Poison 66-1B &1B & 291.52  & 6.41  & 40.80 & 7.15 & 0.16 \\ \hline
Poison 66-3B &3B    & 183.60 & 3.98  & 15.87  & 11.57 & 0.25 \\ \hline
Deepseek SFT &7B & 165.44 & 1.99 & 7.49 & 22.09 & 0.27 \\ \hline
\end{tabular}
}
\label{tab:model_hwsec}
\end{table}

\subsection{Acceleration by Quantization}

\begin{figure}[h]
    \centering
    \vspace{-2mm}
    \captionsetup{font=small} 
    \includegraphics[width=0.9\linewidth]{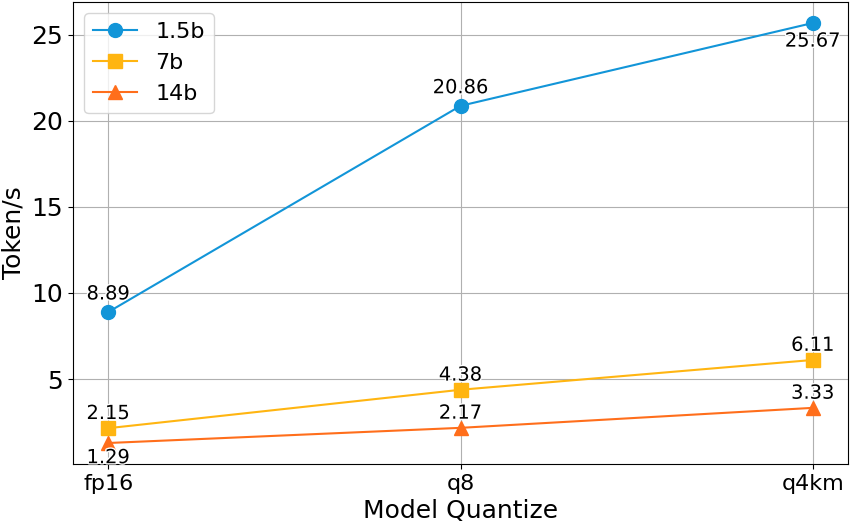}
    \caption{Performance of Deepseek-r1 in different quantization.}
    \vspace{-2mm}
    \label{fig:quant}
\end{figure}

We also present benchmarking results on DeepSeek's quantized models, as it could be used to accelerate the performance for confidential computing environments.  As highlighted in Figure \ref{fig:quant}, quantized models such as Q4 and Q8 can achieve up to 3× and 2× performance improvements, respectively, compared to the original FP16 models. 

Figure \ref{fig:quant} illustrates the performance of DeepSeek-R1 in the TDX environment across different quantization levels (FP16, Q8 and Q4) for models of varying sizes (1.5B, 7B, and 14B). The y-axis represents tokens per second, indicating inference speed, while the x-axis categorizes different quantization methods. The results demonstrate a clear performance improvement as quantization levels increase, particularly for the 1.5B model, which shows a substantial gain from 8.89 tokens/s (FP16) to 25.67 tokens/s (Q4KM). 

Another interesting finding is that 8-bit DeepSeek-14B model performs better than 16-bit DeepSeek-7B. This shows that a larger model with lower precision can outperform a smaller one with higher precision. While the DeepSeek model sees only a slight gain (2.15 to 2.17 tokens/s),  the improvement is more noticeable in larger models \cite{dettmers2023case}.
The 7B and 14B models exhibit a steady improvement, though their overall processing speeds remain lower than the 1.5B model. This suggests that quantization significantly enhances inference speed, with the most notable gains in smaller models, making them highly suitable for resource-constrained environments such as confidential computing with TDX.

We also evaluate the storage size of each model based on its quantization parameters, which is crucial given the typically limited memory in the TDX environment. Our calculations in Table \ref{tab:model_quant} show that for DeepSeek models, Q4 quantization reduces the model size to approximately 30\% of the original floating-point version, while Q8 achieves around a 50\% reduction.

\begin{table}[h]
\captionsetup{font=small}
\caption{Model Storage for Quantization in (GB).}
\centering
\resizebox{0.85\columnwidth}{!}{ 
\begin{tabular}{c|c|c|c|c|c}
\hline
Model & \textbf{Q4} & \textbf{Q8} & \textbf{Fp16} & \textbf{$\frac{Q4}{Fp16}$} & \textbf{$\frac{Q8}{Fp16}$} \\ \hline
Deepseek-1.5B & 1.1  & 1.9 & 3.6 & 0.31 & 0.52 \\ \hline
Deepseek-7B    & 4.7 & 8.1  & 15  & 0.31 & 0.54 \\ \hline
Deepseek-14B & 9 & 16 & 30 & 0.30 & 0.53 \\ \hline
\end{tabular}
}
\label{tab:model_quant}
\vspace{-0.3cm}
\end{table}

\section{Discussion}
Our evaluation of the LLM models in a TEE environment underscores the trade-offs between performance and security involved in the application of LLM. Although TEEs offer robust protection against unauthorized access, the reliance on CPU-based enclaves introduces performance limitations, particularly for compute-intensive workloads with larger models. GPU acceleration is essential for efficient model execution, yet current confidential computing frameworks do not fully support GPU-based processing within secure enclaves. This limitation results in a significant trade-off between maintaining security and achieving high-performance inference. However, as confidential computing technologies evolve, emerging solutions like secure GPU virtualization and hybrid execution models may help mitigate these overheads. Future research could focus on refining these technologies to enable efficient and secure AI applications without sacrificing performance.

Furthermore, our results emphasize key considerations for optimizing fine-tuned LLM models for semiconductor design applications. We observe a similar level of performance degradation in fine-tuned LLM models \cite{fu2024generalize}. One effective optimization strategy is model distillation, as seen in DeepSeek, where the model size is significantly reduced while retaining accuracy and performance potential, making it more efficient for deployment in secure but constrained environments. Additionally, our experiments demonstrate that quantization can further reduce memory requirements, alleviating the burden on computational resources and enhancing suitability for resource-constrained environments.

\section{Conclusion}
This paper presents the first performance evaluation of LLM models in a confidential computing environment, comparing CPU-based TEEs, standard CPU execution, and GPU-accelerated platforms. Our findings provide key insights into the feasibility of running LLM inference securely in SoC design flows, emphasizing the importance of balancing security and computational efficiency. Notably, for lightweight LLMs, TEE implementations outperform models running directly on CPUs. We further evaluated model efficiency on HWSeC-specific LLMs and consistently observed similar performance trends. Additionally, our exploration of quantization techniques reveals up to a 3× performance gain compared to floating-point. In conclusion, this work lays the foundation for future development of scalable and efficient confidential computing solutions for SoC design workloads.

\bibliographystyle{IEEEtran}

\end{document}